\documentclass{AISB2008}
\usepackage{times}
\usepackage{fancyvrb}
\usepackage{graphicx}
\usepackage{latexsym}
\usepackage{hyperref}

\begin{document}

\title{Computing as compression: the SP theory of intelligence}

\author{J Gerard Wolff\institute{CognitionResearch.org, UK, email: jgw@cognitionresearch.org}}

\maketitle
\bibliographystyle{AISB2008}

\begin{abstract}

This paper provides an overview of the {\em SP theory of intelligence} and its central idea that artificial intelligence, mainstream computing, and much of human perception and cognition, may be understood as information compression.

     The background and origins of the SP theory are described, and the main elements of the theory, including the key concept of {\em multiple alignment}, borrowed from bioinformatics but with important differences. Associated with the SP theory is the idea that {\em redundancy} in information may be understood as repetition of patterns, that compression of information may be achieved via the matching and unification (merging) of patterns, and that computing and information compression are both fundamentally probabilistic. It appears that the SP system is Turing-equivalent in the sense that anything that may be computed with a Turing machine may, in principle, also be computed with an SP machine.

     One of the main strengths of the SP theory and the multiple alignment concept is in modelling concepts and phenomena in artificial intelligence. Within that area, the SP theory provides a simple but versatile means of representing different kinds of knowledge, it can model both the parsing and production of natural language, with potential for the understanding and translation of natural languages, it has strengths in pattern recognition, with potential in computer vision, it can model several kinds of reasoning, and it has capabilities in planning, problem solving, and unsupervised learning.

     The paper includes two examples showing how alternative parsings of an ambiguous sentence may be modelled as multiple alignments, and another example showing how the concept of multiple alignment may be applied in medical diagnosis.

\end{abstract}

\section{INTRODUCTION}

Since about 1987, I have been developing the idea that artificial intelligence, mainstream computing, and much of human perception and cognition, may be understood as information compression. An early version of the idea, as applied to computing, is described in \cite{wolff_1990}. Since then, progressively more refined versions of the {\em SP theory of intelligence} have been described in several peer-reviewed articles and, in some detail, in a book \cite{wolff_2006}.\footnote{Bibliographic details of relevant publications may be found via links from \url{www.cognitionresearch.org/sp.htm}.}

The main aim in this paper is to to provide an overview of the theory, with the main focus on computing, and to describe some of the associated thinking.

\section{ORIGINS AND BACKGROUND}

The SP theory has grown out of four main strands of work:

\begin{itemize}

\item A body of research, pioneered by Fred Attneave (eg, \cite{attneave_1954}), Horace Barlow (eg \cite{barlow_1959,barlow_1969}), and others, showing that many aspects of the workings of brains and nervous systems may be understood as compression of information.\footnote{Apart from direct evidence for the importance of information compression in the workings of brains and nervous systems, we would expect information compression to have been favoured by natural selection because it can facilitate economies in the storage of information, economies in the processing and transmission of information, corresponding economies in energy demands (the brain is 2\% of total body weight but it demands 20\% of our resting metabolic rate), and perhaps most importantly, it provides the key to the inductive prediction of the future from the past (Section \ref{compression_probability}).}

\item My own research, developing models of language learning, where the importance of information compression became increasingly clear (see, for example, \cite{wolff_1988}).\footnote{Details of this and other relevant publications may be found via \href{http://www.cognitionresearch.org/lang\_learn.html}{www.cognitionresearch.org/lang\_learn.html}.}

\item Research on principles of `minimum length encoding', pioneered by Ray Solomonoff (eg, \cite{solomonoff_1964}), Chris Wallace (eg, \cite{wallace_boulton_1968}), Jorma Rissanen (eg, \cite{rissanen_1978}), and others.

\item Several observations that suggest that information compression has a key role in computing, mathematics, and logic (\cite[Chapters 2 and 10]{wolff_2006}), some of which are outlined in Section \ref{sp_conventional_computing}, below.

\end{itemize}

\section{OUTLINE OF THE SP THEORY}\label{sp_outline}

The main elements of the SP theory are:

\begin{itemize}

\item The theory is conceived as an abstract system that, like a brain, may receive `New' information via its senses and store some or all of it as `Old' information.

\item All New and Old information is expressed as arrays of atomic symbols ({\em patterns}) in one or two dimensions.\footnote{So far, the main emphasis in the development of the theory has been on one-dimensional patterns, as described in this paper. But there is clear potential to generalise the theory for patterns in two dimensions, with potential applications in, for example, computer vision.}

\item The system is designed for the unsupervised learning of Old patterns by compression of New patterns.

\item An important part of this process is, where possible, the economical encoding of New patterns in terms of Old patterns. This may be seen to achieve such things as pattern recognition, parsing or understanding of natural language, or other kinds of interpretation of incoming information in terms of stored knowledge, including several kinds of reasoning.

\item Compression of information is achieved via the matching and unification (merging) of patterns, with an improved version of dynamic programming \cite[Appendix A]{wolff_2006} providing flexibility in matching, and with key roles for the frequency of occurrence of patterns, and their sizes.

\item The concept of {\em multiple alignment}, outlined in Section \ref{multiple_alignment}, is a powerful central idea, similar to the concept of multiple alignment in bioinformatics but with important differences.

\item Owing to the intimate connection between information compression and concepts of probability (Section \ref{compression_probability}), it is relatively straightforward for the SP system to calculate probabilities for inferences made by the system, and probabilities for parsings, recognition of patterns, and so on.

\item In developing the theory, I have tried to take advantage of what is known about the psychological and neurophysiological aspects of human perception and cognition, and to ensure that the theory is compatible with such knowledge. The way the SP concepts may be realised with neurons ({\em SP-neural}) is discussed in \cite[Chapter 11]{wolff_2006}.

\end{itemize}

These ideas are currently realised in the form of two computer models, SP62 and SP70, described in outline below. These models may be seen as first versions of the proposed {\em SP machine}, an expression of the SP theory and a means for it to be applied.

\subsection{The multiple alignment concept}\label{multiple_alignment}

The multiple alignment concept in the SP theory has been adapted from a similar concept in bioinformatics, where it means a process of arranging, in rows or columns, two or more DNA sequences or amino-acid sequences so that matching symbols---as many as possible---are aligned orthogonally in columns or rows.

The main difference between the two concepts is that, in bioinformatics, all sequences have the same status, whereas in the SP theory, the system attempts to create a multiple alignment which enables one New pattern (sometimes more) to be encoded economically in terms of one or more Old patterns.

\begin{figure*}[!hbt]
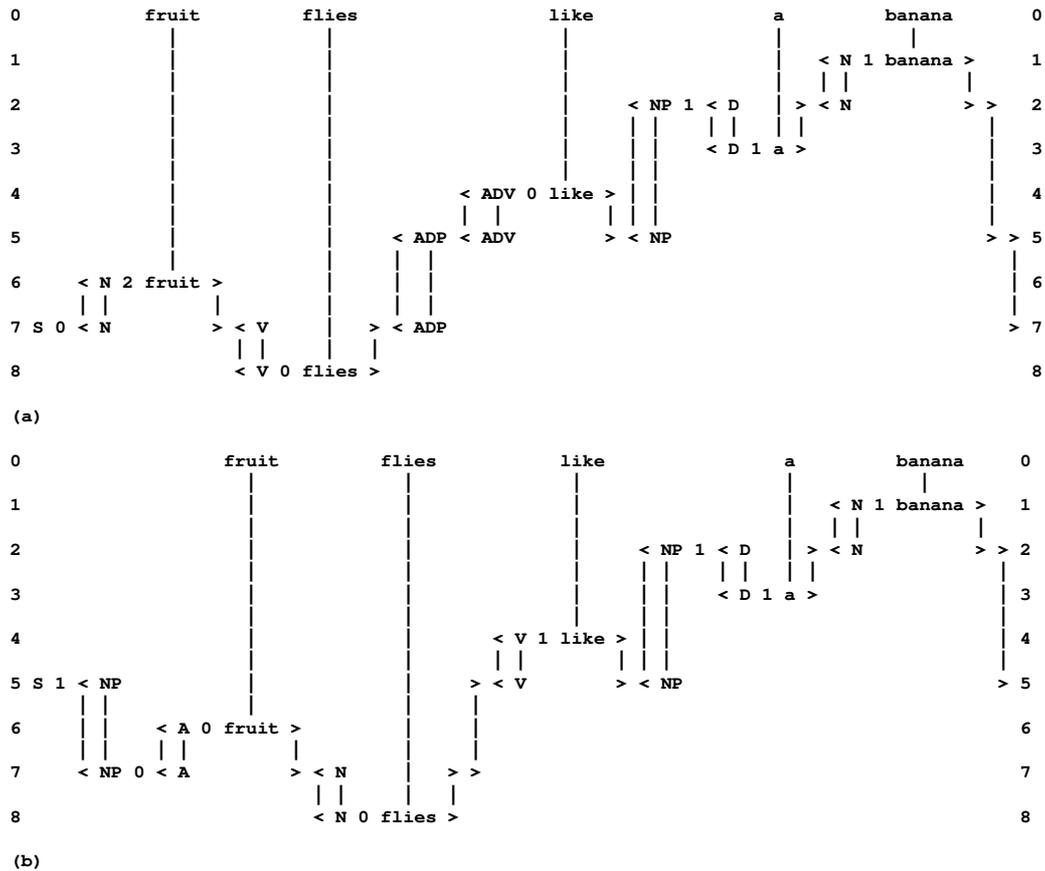

\fontsize{07.00pt}{08.40pt}
\centering
{\bf
\begin{BVerbatim}
0           fruit         flies                 like                a         banana       0
              |             |                    |                  |           |
1             |             |                    |                  |   < N 1 banana >     1
              |             |                    |                  |   | |          |
2             |             |                    |     < NP 1 < D   | > < N          > >   2
              |             |                    |     | |    | |   | |                |
3             |             |                    |     | |    < D 1 a >                |   3
              |             |                    |     | |                             |
4             |             |           < ADV 0 like > | |                             |   4
              |             |           |  |         | | |                             |
5             |             |     < ADP < ADV        > < NP                            > > 5
              |             |     |  |                                                   |
6     < N 2 fruit >         |     |  |                                                   | 6
      | |         |         |     |  |                                                   |
7 S 0 < N         > < V     |   > < ADP                                                  > 7
                    | |     |   |
8                   < V 0 flies >                                                          8

(a)

0                  fruit         flies           like                a         banana     0
                     |             |              |                  |           |
1                    |             |              |                  |   < N 1 banana >   1
                     |             |              |                  |   | |          |
2                    |             |              |     < NP 1 < D   | > < N          > > 2
                     |             |              |     | |    | |   | |                |
3                    |             |              |     | |    < D 1 a >                | 3
                     |             |              |     | |                             |
4                    |             |       < V 1 like > | |                             | 4
                     |             |       | |        | | |                             |
5 S 1 < NP           |             |     > < V        > < NP                            > 5
      | |            |             |     |
6     | |    < A 0 fruit >         |     |                                                6
      | |    | |         |         |     |
7     < NP 0 < A         > < N     |   > >                                                7
                           | |     |   |
8                          < N 0 flies >                                                  8

(b)
\end{BVerbatim}
}
\caption{The two best multiple alignments found by a near-identical precursor of the SP62 computer model with Old patterns representing grammatical rules (rows 1 to 8 in (a) and (b)) and the ambiguous sentence ‘fruit flies like a banana’ as a `New' pattern (row 0 in each multiple alignment). Reproduced from Figure 5.1 in \protect\cite{wolff_2006}, with permission.}
\label{fruit_flies}
\end{figure*}

As an illustration of the concept, Figure \ref{fruit_flies} shows two multiple alignments which are, in effect, two alternative parsings of the ambiguous sentence `Fruit flies like a banana'.\footnote{This sentence is the second part of {\em Time flies like an arrow. Fruit flies like a banana.}, attributed to Groucho Marx.}

These two multiple alignments are the best ones found by the SP61 computer model of the SP system\footnote{SP61 is a slightly earlier precursor of SP62, and very similar to it.}, with a set of Old patterns representing grammatical rules (including words and their grammatical categories) and a New pattern representing the sentence to be parsed. Here, `best' means that these two multiple alignments achieve the greatest degree of compression of the New patterns via its encoding in terms the Old patterns. More detail may be found in \cite[Section 3.5]{wolff_2006}.

Although this example does not illustrate the point, it is pertinent to mention that the process of forming multiple alignments is robust in the face of errors. Plausible multiple alignments may be formed even when the New pattern or the Old patterns, or both, contain errors of omission, commission, or substitution.

\subsection{Computer models}

The SP theory is currently expressed in the SP70 computer model, and a subset of it---SP62---which lacks any ability in learning.

At the heart of the SP models is a process for finding good full or partial matches between patterns \cite[Appendix A]{wolff_2006}, somewhat like the WinMerge utility for finding similarities and differences between files, or standard `dynamic programming' methods for the alignment of sequences. The main difference between the SP process and others, is that the former can deliver several alternative matches between patterns, while WinMerge and standard methods deliver one `best' result.

Multiple alignments are built in stages, with pairwise matching and merging of patterns, with merged patterns from any stage being carried forward to later stages, and with a weeding out, at all stages, of low-scoring multiple alignments. This is broadly similar to some programs for the creation of multiple alignments in bioinformatics.

In the SP70 model, there are additional processes of deriving Old patterns from multiple alignments, evaluating sets of newly-created Old patterns in terms of their effectiveness for the economical encoding of the New information, and the weeding out low-scoring sets.

Because of the way each model searches for a global optimum, it does not depend on the presence or absence of any particular feature or combination of features. Up to a point, plausible results may be obtained in the face of errors of omission, commission and substitution in the data.

More detail may be found in \cite[Sections 3.9, 3.10, and 9.2]{wolff_2006}.

\subsection{Information compression and the matching and unification of patterns}\label{compression_matching_unification}

A central idea in the SP theory and the multiple alignment concept is that {\em redundancy} in information may be understood as repetition of patterns, and that information compression may be achieved by finding patterns that match each other and merging or `unifying' patterns that are the same. For the sake of brevity, this kind of matching and unification of patterns will be referred to as `MUP'.

MUP is fairly easy to see in widely-used `zip' programs for information compression, based on the LZW algorithm or related algorithms. But such processes are invisible or hard to see in compression techniques such as arithmetic coding or wavelet compression.

Although the point has not been argued in detail, I believe it is likely that MUP is fundamental, not only in zip programs, but also in compression techniques with a mathematical orientation:

\begin{itemize}

\item In \cite[Chapter 10]{wolff_2006}, I have argued that much of mathematics (and logic) may be understood in terms of information compression via multiple alignment, including MUP.

\item If that argument is accepted then, since mathematics is prominent in compression techniques such as arithmetic coding and wavelet compression, we may infer that MUP is fundamental in compression techniques of that kind.

\end{itemize}

\subsubsection{The matching and unification of patterns, heuristic search, and computational complexity}\label{heuristic_search}

In considering MUP, it not hard to see that, for any body of information $I$, except very small examples, there is a huge number of alternative ways in which patterns may be matched against each other, and there will normally be many alternative ways in which patterns may be unified \cite[Section 2.2.8.4]{wolff_2006}.

As indicated earlier, the focus of interest is normally on matches between patterns that yield relatively high levels of compression. Since it is not normally possible to make an exhaustive search of the space of alternative matches, the SP computer models rely on the kinds of heuristic techniques that are familiar in other AI applications, reducing the size of the search space by pruning the search tree at appropriate points. Current models allow the user to apply various constraints on searching (such as the exclusion of partial matches between patterns), and to choose any size of search space from large (slow but thorough) through to small (quick and dirty).

In its ideal form, with exhaustive search, and with a realistically large size for $I$, MUP is not tractable. But with the use of heuristic techniques, MUP becomes quite practical, with time complexity in a serial processing environment estimated to be O$(n \cdot m)$, where $n$ is the number of atomic symbols in a given pattern and $m$ is the number of atomic symbols in $I$ \cite[Appendix A.4]{wolff_2006}. In a parallel-processing environment, the time complexity has been estimated to be O$(n)$ ({\em ibid.}).

\subsection{Information compression, prediction, and probability}\label{compression_probability}

It is widely recognised that there is a close connection between information compression and concepts of prediction and probability (see, for example, \cite{li_vitanyi_2009}). In terms of the SP theory, that close connection makes good sense:

\begin{itemize}

\item The amount of compression that can be achieved via MUP depends directly on the sizes of patterns that are unified and their frequencies. The patterns that yield relatively high levels of compression are also those that provide a good basis for inductive prediction.\footnote{In this connection, it appears that the sizes of patterns are as important as their frequencies.}

\item Partial matches between patterns provide the basis for specific predictions: if we are going out and we see black clouds then, knowing the association between black clouds and rain, we may decide to take an umbrella.

\end{itemize}

All of this is fundamentally probabilistic:

\begin{itemize}

\item As mentioned above, the frequency of occurrence of patterns is a key variable in the search for unifications that yield high levels of compression.

\item Because it is not normally possible to achieve ideal solutions or even to know whether or not such a solution has been found (Section \ref{heuristic_search}), there will be corresponding uncertainties.

\item For every multiple alignment that is created by the system, and for any inferences that may be drawn, there is an associated probability. These probabilities can be calculated by the SP computer models.

\end{itemize}

There is more on this topic in Section \ref{computing_probabilities}.

\section{THE SP THEORY AND CONCEPTS OF `COMPUTING'}\label{sp_computing}

\subsection{The SP theory and the Turing model of computing}\label{sp_turing}

In \cite[Chapter 4]{wolff_2006}, I have argued that the SP system is equivalent to a universal Turing machine \cite{turing_1936}, in the sense that anything that may be computed with a Turing machine may, in principle, also be computed with an SP machine. The `in principle' qualification is necessary because the SP theory is still not fully mature and there are still some weaknesses in the SP computer models.

The gist of the argument is that the operation of a Post canonical system \cite{post_1943} may be understood in terms of the SP theory and, since it is accepted that the Post canonical system is equivalent to the Turing machine (as a computational system), the Turing machine may also be understood in terms of the SP theory.

The thread running through all three models of computing is the matching and unification of patterns. This is, of course, a prominent feature of the SP theory. Although it is not formally recognised in the Post canonical system or the Turing machine, it is relatively clear to see in the Post canonical system and it can also be seen to operate in the state transition tables of the Turing machine.

Is there anything to choose between these three models? Isn't the SP theory just another model of computing to go alongside earlier models such as the Turing and Post models, `lamda calculus' \cite{rosser_1984}, `recursive function' \cite{kleene_1936}, and `normal algorithm' \cite{markov_nagorny_1988}?

In answer to those questions, the main differences between the SP theory and earlier theories of computing are:

\begin{itemize}

\item It has a lot more to say about the nature of `intelligence' than other theories of computing (see Section \ref{sp_intelligence}).

\item Unlike earlier theories, it is founded on principles of information compression via the matching and unification of patterns, and it includes mechanisms for building multiple alignments and for heuristic search that are not present in any of the other models.

\item Although the SP system is more complex than the Turing model of computing, it can mean substantial reductions in the overall complexity of computing systems, bearing in mind that software is as much part of a computing system as is the hardware (\cite[Section 4.4.2]{wolff_2006}, \cite{sp_benefits_apps}). The reasoning is that, by providing MUP mechanisms in the core of SP system, there is less need to provide those mechanisms in software and, in particular, there is less need to repeat those mechanisms again and again in different software applications.

\end{itemize}

\subsection{The SP theory and conventional computing}\label{sp_conventional_computing}

Although it may seem counter-intuitive to suppose that information compression has any significant role in conventional computing as we know it today, there is evidence in support of that idea, as outlined in the following subsections.

\subsubsection{Information compression via the matching and unification of patterns}

The matching of patterns is widespread in conventional computing systems and in most cases there is at least an implicit unification of the patterns that match each other. Here are some examples:

\begin{itemize}

\item {\em Accessing information in computer memory}. The process of accessing an item of information in computer memory means finding a match between the address of the item as it is known within the CPU and the address of the item in computer memory. Although the process is normally described in terms of the operations of logic circuits, that should not obscure the fact that it is a process of finding a match between two copies of the relevant address.

\item {\em De-referencing of names}. Names are widely used in conventional computing. Examples include the names of functions, procedures or sub-routines, names for objects and classes in object-oriented programs, names for tables, records, and fields in databases, names of files and directories, names of variables, arrays and other data structures, and labels for program statements (for use with the now-shunned `go to' statements). The de-referencing of any such name---finding what it represents---means finding a match, with an implicit unification, between the name as a reference and the name on the structure that is to be retrieved.

\item {\em Information retrieval}. The `query-by-example' technique of retrieving information from a database means searching for a good match (full or partial) between the query and zero or more records in the database, with implicit unification where matches are found. In a similar way, searching for information on the internet means searching for good full or partial matches between a query pattern and zero or more web pages.

\end{itemize}

\subsubsection{Information compression via chunking-with-codes}\label{chunking_with_codes}

The `chunking-with-codes' technique for the compression of information means identifying a relatively large `chunk' of information that occurs two or more times in a body of information, giving it a relatively short name or `code', and then using the name instead the chunk in all but one places where the chunk occurs.

Perhaps the most obvious example of this idea in conventional computing is the use of named functions, procedures or sub-routines in computer programs. The function or procedure may be seen as a chunk of information which is defined in one part of a given program and accessed via its name from other parts of the program. Unless the given function is used only once, this technique will normally mean useful savings in the sizes of programs. It will also facilitate the editing of computer programs and eliminate the risk of introducing inconsistencies between different instances of a given function.

Similar things can be said about most of the other kinds of names mentioned above in connection with `de-referencing of names'.

\subsubsection{Information compression via run-length coding}\label{run_length_coding}

In sequential information, the `run-length coding' technique for the compression of information may be applied wherever something is repeated two or more times in a contiguous sequence. In that case, multiple instances may be reduced to one, with some kind of indication that it repeats, something like `a b c (10)' (showing that the pattern `a b c' repeats 10 times) or `a b c (*)' (showing that the pattern repeats but without specifying the number of repetitions).

In computer programs, this kind of technique can be seen in iterations (eg, {\em repeat ... until}, {\em while ... do}, or {\em for ... do}) and also in recursive functions such as:

\begin{center}
\begin{BVerbatim}
long factorial(int x)
{
     if (x == 1) return(1) ;
     return(x * factorial(x - 1)) ;
}.
\end{BVerbatim}
\end{center}

\noindent The use of iteration or recursion can avoid a lot of space-wasting redundancy in computer programs.

\subsubsection{Information compression via schema-plus-correction}\label{schema_plus_correction}

The `schema-plus-correction' technique for information compression may be applied where a pattern is repeated but with variations from one occurrence to another. For example, a six-course menu in a restaurant may have the general form `Appetiser, S, sorbet, M, P, coffee and mints', with choices at the points marked `S' (starter), `M' (main course), and `P' (pudding). Then a particular meal may be encoded economically as something like `Menu1 (3)(5)(1)', where the digits determine the choices of starter, main course, and pudding.

In a computer program, any function or procedure that has parameters may be seen as a schema, where the parameters serve to determine choices within the schema---and where those choices are normally expressed in the form of conditional statements.

In object-oriented design (next section), a class may also be seen as a schema for particular objects, with the details of each object determined via parameters.

\subsubsection{Information compression via object-oriented design}\label{oo_design}

In terms of information compression, object-oriented design appears to be significant for two main reasons:

\begin{itemize}

\item {\em Economies in human perception and cognition}. Arguably, we see the world in terms of discrete objects because each such object is a recurrent constellation of features (a `chunk') that enables us to perceive and understand things in an economical way \cite[Section 2.3.2]{wolff_2006}. Likewise, there are huge economies to be made in recognising things in terms of classes ({\em ibid.}).

\item {\em Economies in software design}. By modelling software on the objects and classes that people know, we not only create programs that are easy for people to understand but we take advantage of what are normally very substantial economies in the way that people understand things. As already indicated, software objects may be seen as examples of the chunking-with-codes technique for compressing information and object-oriented classes may be seen as examples of schema-plus-correction.

\end{itemize}

\subsubsection{Information compression in mathematics and logic}

As indicated earlier, similar things may be said about mathematics and logic \cite[Chapter 10]{wolff_2006}. It appears that much of mathematics and logic may be understood in terms of the kinds of compression techniques that have been mentioned: chunking-with-codes, run-length coding, and schema-plus-correction.

As an example of the power of mathematics to compress information, Newton's equation that relates the distance travelled by a falling object to the time since it began to fall ($s = gt^2 / 2$) is very much more compact than any realistically-large table of those distances and times.

\subsection{Computing and probabilities}\label{computing_probabilities}

As indicated in Section \ref{compression_probability}, there is an intimate connection between information compression and concepts of probability, and the SP system is fundamentally probabilistic. This implies that computing is fundamentally probabilistic.

That may seem like a strange conclusion in view of the clockwork certainties that we associate with the operation of ordinary computers and the workings of mathematics and logic. There are at least three answers to that apparent contradiction:

\begin{itemize}

\item It appears that computing, mathematics and logic are more probabilistic than our ordinary experience of them might suggest. Gregory Chaitin has written: ``I have recently been able to take a further step along the path laid out by G\"{o}del and Turing. By translating a particular computer program into an algebraic equation of a type that was familiar even to the ancient Greeks, I have shown that there is randomness in the branch of pure mathematics known as number theory. My work indicates that---to borrow Einstein’s metaphor---God sometimes plays dice with whole numbers.'' \cite[p. 80]{chaitin_1988}.

\item The SP system may imitate the clockwork nature of ordinary computers by delivering probabilities of 0 and 1. This can happen with certain kinds of data, or tight constraints on the process of searching the abstract space of alternative matches, or both those things.

\item It seems likely that the all-or-nothing character of conventional computers has its origins in the low computational power of early computers. In those days, it was necessary to apply tight constraints on the process of searching for matches between patterns. Otherwise, the computational demands would have been overwhelming. Similar things may be said about the origins of mathematics and logic, which have been developed for centuries without the benefit of any computational machine, except very simple and low-powered devices. Now that it is technically feasible to apply large amounts of computational power, constraints on searching may be relaxed.

\end{itemize}

\section{THE SP THEORY AND THE NATURE OF INTELLIGENCE}\label{sp_intelligence}

Although Alan Turing envisaged that computers might become intelligent \cite{turing_1950}, the Turing theory, in itself, does not tell us how! Plugging that gap has been an important motivation in the development of the SP theory. As it stands, it is certainly not a comprehensive answer but, as was mentioned in Section \ref{sp_turing}, and amplified here, it does have a lot more to say about the nature of intelligence than earlier theories of computing.

The most comprehensive account of these aspects of the SP theory is in \cite{wolff_2006}. In brief:

\begin{itemize}

\item {\em Representation of knowledge and information retrieval}. Despite the simplicity of representing knowledge with patterns, the way they are processed within the multiple alignment framework gives them the versatility to represent several kinds of knowledge, including grammars for natural languages (next bullet point), class hierarchies, part-whole hierarchies, decision networks and trees, relational tuples, if-then rules, associations of medical signs and symptoms (Section \ref{medical_diagnosis_section}), causal relations, and more. One universal format for knowledge and one universal framework for processing means that different kinds of knowledge may be combined flexibly and seamlessly according to need. The SP system provides for the retrieval of information from a knowledge base in the manner of query-by-example, and has potential to support the development of query languages, if required. The system may serve as an intelligent database that also supports the use of traditional data models---but with advantages compared with existing systems \cite{wolff_sp_intelligent_database}.

\item {\em Natural language processing}. Grammatical rules, including words and their grammatical categories, may be represented with SP patterns. As we have seen (Figure \ref{fruit_flies}) the parsing of natural language may be modelled via the building of multiple alignments. The same is true of the production of natural language. The framework provides an elegant means of representing discontinuous dependencies in syntax, including overlapping dependencies such as number dependencies and gender dependencies in languages like French. As indicated in the previous item, the system may also model non-syntactic `semantic' structures and, because there is one simple format for different kinds of knowledge, the system facilitates the seamless integration of syntax with semantics---with a consequent potential for the understanding of natural languages and interlingua-based translations amongst languages. The system is robust in the face of errors of omission, commission or substitution in sentences to be analysed, or stored linguistic knowledge, or both. The importance of context in the processing of language is accommodated in the way the system searches for a global best match for patterns: any pattern or partial pattern may be a context for any other.

\item {\em Pattern recognition and computer vision}. Thanks largely to the versatility of the multiple alignment concept, the SP system provides a powerful framework for pattern recognition. It can model pattern recognition at multiple levels of abstraction, it provides for cross-classification and the integration of class-inclusion relations with part-whole hierarchies, and it facilitates the seamless integration of pattern recognition with various kinds of reasoning (next bullet point), and other aspects of intelligence. A probability may be calculated for any given classification or any associated inference. As in the processing of natural language, the system is robust in the face of errors of omission, commission or substitution in incoming data, or stored knowledge, or both, and the importance of context in recognition is accommodated in the way the system searches for a global best match for patterns. These ideas appear to have potential in the field of computer vision, as discussed in \cite{sp_vision}.

\item {\em Reasoning}. The SP system can model several kinds of reasoning including one-step `deductive' reasoning, abductive reasoning, reasoning with probabilistic decision networks and decision trees, reasoning with `rules', nonmonotonic reasoning and reasoning with default values, reasoning in Bayesian networks (including `explaining away'), causal diagnosis, and reasoning which is not supported by evidence. Since these several kinds of reasoning all flow from one computational framework (multiple alignment), they may be seen as aspects of one process, working individually or together without inconsistencies or incompatibilities. Plausible lines of reasoning may be achieved, even when relevant information is incomplete. Probabilities of inferences may be calculated, which may, as previously indicated, include extreme values (0 or 1) in the case of logic-like `deductions'.

\item {\em Planning and problem solving}. The SP framework provides a means of planning a route between two places, and, with the translation of geometric patterns into textual form, it can solve the kind of geometric analogy problem that may be seen in some puzzle books and IQ tests \cite[Chapter 8]{wolff_2006}.

\item {\em Unsupervised learning}. The SP70 model can derive a plausible grammar from a set of sentences without supervision or error correction by a `teacher', without the provision of `negative' samples, and without the grading of samples from simple to complex. It thus overcomes restrictions on what can be achieved with some other models of learning and reflects more accurately what is known about how children learn their first language or languages. The model draws on earlier research showing that inductive learning via principles of `minimum length encoding' can lead to the discovery of entities that are psychologically natural---such as words in natural languages \cite{wolff_1977}. As it stands now, the model is not able to derive intermediate levels of abstraction or discontinuous dependencies in data, but those problems appear to be soluble.

\end{itemize}

\subsection{A second example of multiple alignment: medical diagnosis}\label{medical_diagnosis_section}

To illustrate some of the versatility of the multiple alignment concept, Figure \ref{medical_diagnosis_figure} shows how it may be applied to medical diagnosis \cite{wolff_medical_diagnosis}.\footnote{As a matter of detail, the patterns in this multiple alignment are arranged in columns instead of rows, so that the multiple alignment can be fitted more neatly on to a page.} This is the best multiple alignment found by SP62 with a set of Old patterns that provide information about diseases and a set of New patterns that describe the symptoms of an imaginary patient, `John Smith'.

\begin{figure*}[!hbt]
\fontsize{06.70pt}{08.04pt}
\centering
{\bf
\begin{BVerbatim}
0                1                    2                    3              4                5

                 <disease> ---------- <disease> ---------- <disease> ---- <disease>
                                      flu
                 : ------------------ :
<patient> ------ <patient>
John_Smith
</patient> ----- </patient>
                 <dname> ------------ <dname>
                                      Influenza
                 </dname> ----------- </dname>
                 <R1> --------------- <R1>
                                      flu_symptoms ------- flu_symptoms
                 </R1> -------------- </R1>
                 <R2> ------------------------------------ <R2>
                                                           fever -------- fever
                 </R2> ----------------------------------- </R2>
<appetite> ----- <appetite> --------- <appetite>
poor                                  normal
</appetite> ---- </appetite> -------- </appetite>
<breathing> ---- <breathing> -------------------------------------------- <breathing>
rapid ------------------------------------------------------------------- rapid
</breathing> --- </breathing> ------------------------------------------- </breathing>
                 <chest> ------------ <chest>
                                      normal
                 </chest> ----------- </chest>
<chills> ------- <chills> -------------------------------- <chills>
yes ------------------------------------------------------ yes
</chills> ------ </chills> ------------------------------- </chills>
                 <cough> --------------------------------- <cough>
                                                           yes
                 </cough> -------------------------------- </cough>
                 <diarrhoea> -------- <diarrhoea>
                                      no
                 </diarrhoea> ------- </diarrhoea>
<face> --------- <face> ------------------------------------------------- <face>
flushed ----------------------------------------------------------------- flushed
</face> -------- </face> ------------------------------------------------ </face>
<fatigue> ------ <fatigue> ---------- <fatigue>
yes                                   no
</fatigue> ----- </fatigue> --------- </fatigue>
                 <headache> ------------------------------ <headache>
                                                           yes
                 </headache> ----------------------------- </headache>
<lymph_nodes> -- <lymph_nodes> ------ <lymph_nodes>
normal ------------------------------ normal
</lymph_nodes> - </lymph_nodes> ----- </lymph_nodes>
<malaise> ------ <malaise> ---------- <malaise>
no ---------------------------------- no
</malaise> ----- </malaise> --------- </malaise>
<muscles> ------ <muscles> ------------------------------- <muscles>
aching --------------------------------------------------- aching
</muscles> ----- </muscles> ------------------------------ </muscles>
<nose> --------- <nose> ---------------------------------- <nose>
runny ---------------------------------------------------- runny
</nose> -------- </nose> --------------------------------- </nose>
                 <skin> ------------- <skin>
                                      normal
                 </skin> ------------ </skin>
<temperature> -- <temperature> ------------------------------------------ <temperature>
                                                                          <t1> ----------- <t1>
38-39 ------------------------------------------------------------------------------------ 38-39
                                                                          </t1> ---------- </t1>
</temperature> - </temperature> ----------------------------------------- </temperature>
<throat> ------- <throat> -------------------------------- <throat>
sore ----------------------------------------------------- sore
</throat> ------ </throat> ------------------------------- </throat>
                 <weight_change> ---- <weight_change>
                                      no
                 </weight_change> --- </weight_change>
                 <causative_agent> -- <causative_agent>
                                      flu_virus
                 </causative_agent> - </causative_agent>
                 <treatment> -------- <treatment>
                                      flu_treatment
                 </treatment> ------- </treatment>
                 </disease> --------- </disease> --------- </disease> --- </disease>

0                1                    2                    3              4                5
\end{BVerbatim}
}
\caption{The best multiple alignment found by SP62 with a set of patterns in New describing the symptoms of the patient `John Smith' and a set of patterns in Old describing a range of different diseases and named clusters of symptoms. In the multiple alignment, New patterns are in column 0 and Old patterns are in columns 1 to 5. The pattern in column 1 serves as a framework for the multiple alignment. Reproduced from Figure 6.11 in \protect\cite{wolff_2006}, with permission.}.
\label{medical_diagnosis_figure}
\end{figure*}

In the example, all the New patterns appear in column 0. They show the name of the patient (`$<$patient$>$ John\_Smith $<$/patient$>$') and his symptoms (`$<$appetite$>$ poor $<$/appetite$>$', `$<$breathing$>$ rapid $<$/breathing$>$', and so on).

The Old patterns in the multiple alignment, one in each of columns 1 to 5, have various roles:

\begin{itemize}

\item {\em Column 1}. This is simply a framework for different aspects of any disease, to facilitate the building of multiple alignments.

\item {\em Column 2}. This shows that the most likely explanation of John Smith's symptoms is that he has influenza.

\item {\em Column 3}. This pattern represents a set of `flu symptoms'. The reason that they are not shown within the main pattern for influenza (column 2) is that the same symptoms can appear in other diseases, most notably smallpox.

\item {\em Column 4}. This pattern shows the symptoms of `fever'. As before, they are shown in a separate pattern because this cluster of symptoms can appear in several different diseases.

\item {\em Column 5}. This pattern, `$<$t1$>$ 38-39 $<$/t1$>$', is, in effect, a `value' for a `variable' in column 4 (`$<$t1$>$ $<$/t1$>$'), which serves to record the patient's temperature.

\end{itemize}

More detail may be found in \cite{wolff_medical_diagnosis} and \cite[Section 6.5]{wolff_2006}.

\section{CONCLUSION}

One of the main strengths of the SP theory is that it has a lot more to say about the nature of intelligence than earlier theories of computing.

A useful step forward in the development of these ideas would be the creation of a version of the SP machine as a high-parallel, open-source, software virtual machine, accessible via the web to researchers everywhere, with a good user interface. This would provide a means for researchers to explore what can be done with the system and to improve it.

% \bibliography{latex_references}
% \input{aisb2013_refs.bbl}

\end{document}